# Adaptive Knowledge-based Multi-Objective Evolutionary Algorithm for Hybrid Flow Shop Scheduling Problems with Multiple Parallel Batch Processing Stages

Feige Liu, Xin Li, Chao Lu, Wenying Gong

*Abstract*—Parallel batch processing machines have extensive applications in the semiconductor manufacturing process. However, the problem models in previous studies regard parallel batch processing as a fixed processing stage in the machining process. This study generalizes the problem model, in which users can arbitrarily set certain stages as parallel batch processing stages according to their needs. A Hybrid Flow Shop Scheduling Problem with Parallel Batch Processing Machines (PBHFSP) is solved in this paper. Furthermore, an Adaptive Knowledge-based Multi-Objective Evolutionary Algorithm (AMOEA/D) is designed to simultaneously optimize both makespan and Total Energy Consumption (TEC). Firstly, a hybrid initialization strategy with heuristic rules based on knowledge of PBHFSP is proposed to generate promising solutions. Secondly, the disjunctive graph model has been established based on the knowledge to find the critical-path of PBHFS. Then, a critical-path based neighborhood search is proposed to enhance the exploitation ability of AMOEA/D. Moreover, the search time is adaptively adjusted based on learning experience from Q-learning and Decay Law. Afterward, to enhance the exploration capability of the algorithm, AMOEA/D designs an improved population updating strategy with a weight vector updating strategy. These strategies rematch individuals with weight vectors, thereby maintaining the diversity of the population. Finally, the proposed algorithm is compared with state-of-the-art algorithms. The experimental results show that the AMOEA/D is superior to the comparison algorithms in solving the PBHFSP.

*Index Terms*—Parallel Batching，Hybrid Flow Shop Problem (HFSP)，Multi-Objective Evolutionary Algorithm Based on Decomposition (MOEA/D).

## I. INTRODUCTION

With the intensification of market competition, manufacturing enterprises are facing pressure to reduce production costs and deliver products to customers in a timely manner. One of the key methods to meet these demands is to increase the processing capabilities of machine equipment, enabling machines to handle operations in batches[1]. Parallel batch processing machines allow multiple jobs to be processed simultaneously, a concept known as parallel batch processing. Parallel batch processing finds applications in various industries such as automotive gear manufacturing[2], chemical production processes in tanks and kilns[3], and healthcare[4]. However, most of the literature on the parallel batch processing is related to the semiconductor industry. Semiconductor manufacturing involves the production of integrated circuits (ICs) on wafers made of silicon or gallium arsenide, a process that is highly complex[5]. Within these intricate processes, some operations are carried out on batch processing machines. A single parallel batch processing machine is applied in burn-in operations in the semiconductor industry that are used to test electronic circuits[6]. In the frontend operations, many process flows have wet cleaning (WET) operations followed by diffusion furnace (FRN) operations. WET and FRN are both batch processes with different maximum batch sizes[7]. Hu et al.[8] cited an example of a batch created using the AM software application. In semiconductor wafer fabrication plants, more than one-third of the operations are carried out on parallel batch processing machines. Moreover, there are continuous batch processing stages in the wafer fabrication process, with wafers following the same stage sequence. However, so far, few studies have focused on the scheduling problems with continuous batch processing stages. Additionally, wafer fabrication imposes highly demanding requirements on the production environment, necessitating cleanroom conditions and extremely complex mechanical equipment. Production is highly energy-intensive, requiring significant electricity consumption[9]. Therefore, it is imperative to study multi batch processing stage scheduling. This paper proposes a Hybrid Flow Shop Scheduling Problem on Parallel Batch Processing Machines (PBHFSP) based on the characteristics of wafer production mentioned above. Based on the Hybrid Flow Shop Scheduling Problem (HFSP), PBHFSP allows any processing stage to be a batch processing stage. All machines in the batch processing stages are parallel batch processing machines. PBHFSP simultaneously optimizes two objectives: makespan and total energy consumption (TEC).

Obviously, PBHFSP is an NP-hard problem[10], and it is difficult to find the optimal solution in a limited time. Designing an algorithm to address the multi batch processing stages scheduling problem, enabling scheduling schemes to maximize enterprise profits, is imperative. However, it is difficult to solve large-scale instances using exact algorithms[11]. MOEA/D is a commonly used evolutionary algorithm for solving multi-objective shop scheduling problems. MOEA/D adopts the method of evolutionary iterative solution set to find a relatively optimal and feasible scheduling solution for the problem. Existing research focuses on improving the search capability of MOEA/D. However, this will reduce the diversity of the population[12]. Therefore, this paper makes improvements to balance the exploitation and exploration capabilities of MOEA/D. An Adaptive Knowledge-based Multi-Objective Evolutionary Algorithm (AMOEA/D) is proposed to solve PBHFSP. Based on the encoding method and the knowledge of PBHFSP, this paper integrates a critical path-based neighborhood structure into AMOEA/D. The range of neighborhood search is adjusted adaptively based on learning experience from Q-learning and Decay Law. Considering the shortcomings of MOEA/D in weight vector selection and maintaining solution diversity, AMOEA/D introduces a population update strategy and weight vector update operations. The main contributions of this paper are as follows:

1) Consider that any stage of HFSP can be a parallel batch processing stage with arbitrary sizes and different capacities of machines.

2) Propose a disjunctive graph based on the knowledge of PBHFSP for associating batch processing stages and discrete processing stages.

3) Propose an adaptive knowledge-based multi-objective evolutionary algorithm to optimize both TEC and makespan objectives.

4) Conduct comprehensive experiments to validate the effectiveness and competitive performance of AMOEA/D.

The rest of this paper is organized as follows: Section II is an overview of literature review. Section III introduces the problem. Section IV provides a detailed description of the AMOEA/D. Section V tests the influence of parameters and the effectiveness of various strategies in the AMOEA/D. Section VI summarizes the conclusions of this article and discusses future development directions.

II. LITERATURE REVIEW

A. Batch processing scheduling

Serial batching and parallel batching are two types of batching. The former processes jobs in the same batch sequentially, while the latter processes jobs in the same batch simultaneously[1]. This paper mainly considers the parallel batch scheduling problem. Due to the widespread application of parallel batch processing in semiconductor production, batch scheduling problems have received widespread attention. For example, Zhou et al.[13] studied batch processors with different capacities and processing speeds based on the original single parallel batch processors. Wang et al.[14] investigated the scheduling problem on parallel batch processing machines with different capacities and non-identical processing powers. Wu et al.[15] studied a scheduling problem of diffusion furnaces in semiconductor fabrication facilities and proposed a dynamic programming model. Pedram et al.[16] considered a non-identical parallel machines batch processing problem with release dates, due dates and variable maintenance activity. Davi et al.[17] developed Iterated Greedy (IG) algorithms combining a random variable neighborhood descent local search procedure, using four neighborhood structures to solve the parallel machine scheduling problem. Xiao et al.[18] obtained an efficient 2-approximation algorithm to solve the problem of scheduling jobs with equal lengths and arbitrary sizes on uniform parallel batch machines with different capacities. All of the above are scheduling problems which only have one processing stage, and the stage is parallel batch processing.

Subsequently, some studies applied parallel batch processing machines to shop scheduling problems. Stefan et al.[19] studied a real-world multi-mode multi-project scheduling problem. There are a total of 16 processing stages in the problem, of which the first and third stages are batch processing stages. Amin et al.[20] considered a Hybrid Flow Shop Problem with parallel batching. In the problem, a machine can be a batch processing machine, and the capacity of the machine is randomly set, and the size of the job is the same. Andrea et al.[21] introduced a hybrid flow shop with set-up time, parallel batching machines with compatible parallel batch families. All machines in this problem is parallel batching machines. Costa et al.[22] proposed the hybrid flow shop with parallel batching, the job belonging to same family can be assigned to the batch and have same processing time. Andy et al.[23] exhibited an enhanced mixed integer programming (MIP) model for a flexible job shop scheduling problem with a parallel batch processing stage. Subsequently, Andy [24] studied the problem of rotating parallel batch processing stages and general flexible job shop scheduling processing stages. Liu et al.[25] investigated a specialized two-stage hybrid flow shop scheduling problem with parallel batching machines considering a job-dependent deteriorating effect and non-identical job sizes simultaneously. Cheng et al.[26] adopted an augmented simulated annealing algorithm to solve a mixed-model assembly job shop scheduling problem with batch transfer. Zeng et al.[27] proposed an auction-based approach with an improved disjunctive graph model for a job shop scheduling problem with parallel batch processing. Wang et al.[28] studied an independent double DQN-based multi-agent reinforcement learning approach for online two-stage hybrid flow shop scheduling with batch machines. Wu et al.[29] propose polynomial-time algorithms based on the observation that the robust single-machine batch scheduling problems are reducible. The aforementioned studies integrate parallel batch processing with hybrid flow shop scheduling problems, treating the batch processing procedure as a specific processing stage in most of their proposed problem models. Compared to models that consider any stage as a potential batch processing stage, these studies lack general applicability. Furthermore, although some studies have considered the possibility of any stage being a batch processing stage[20], they assume that all job sizes are identical. Therefore, it is necessary to define a problem model for combining batch processing and hybrid flow shop scheduling that possesses universal applicability.

## B. MOEA/D in scheduling

MOEA/D is a decomposition-based multi-objective evolutionary algorithm that can also be used to solve shop problems. Wu et al.[30] proposed an improved multi-objective evolutionary algorithm based on decomposition with local search (IMOEA/D-LS). IMOEA/D-LS adopts a collaborative initialization strategy, using multiple crossover mutation operators to improve the search ability of the algorithm. Lucas et al.[31] presented the MOEA/D with updating when required (MOEA/D-UR) that uses a metric that detects improvements so as to determine when to adjust weights. Shao et al.[32] designed an ant colony optimization behavior-based multi-objective evolutionary algorithm based on decomposition. This algorithm is based on the behavior of ant colonies to construct offspring individuals. A multi-strategy dynamic evolution-based improved multi-objective evolutionary algorithm based on decomposition (IMOEA/D) was proposed by Liu et al.[33]. Cao et al.[34] considered a two-stage evolutionary strategy based MOEA/D. The first stage focuses on pushing the solutions into the area of the Pareto front and speeding up its convergence ability, after that, the second stage conducts in the operating solution's diversity and makes the solutions distributed uniformly. The above-mentioned studies have employed various methods to enhance the diversity of solutions generated by MOEA/D. However, few have considered the weight vectors and solution matching. Additionally, existing algorithms are typically designed for solving single batch stage problem or batch processing stage as a special stage in shop scheduling problems. There is currently no research on using MOEA/D to solve the multi-objective parallel batch processing problem. Therefore, it is necessary to propose a multi-objective evolutionary algorithm that can be used to solve PBHFSP.

## III. PROBLEM DESCRIPTION

### A. Problem statement

The PBHFSP can be described as follows. PBHFSP releases the constraint that a machine can only process one job at a time based on the hybrid flow shop scheduling problem. According to the type of machines, the processing stages of jobs are divided into parallel batch processing stages and discrete processing stages. In parallel batch processing stages, the processing of jobs can be done simultaneously under the condition of satisfying machine capacity constraints, while in discrete processing stages, only one job can be processed at a time. Furthermore, the start processing time and release time of operations in the same batch are the same. In more detail, the size of each job is different, but smaller than the capacity of all machines. The capacity of all batch processing machines can also be set to be different. For batch processing machines, the more jobs are processed in a batch, the greater the energy consumption during this period of time. For details, see the third section of Section IV.D below.
In brief, the PBHFSP can be subdivided into three sub-problems, namely, selecting machines for jobs at each stage, determining the batch of operations on the machine, and determining the processing sequence of all batches on machines. The PBHFSP has two optimization objectives: makespan and TEC.

### B. Mathematical Model of PBHFSP

The related notation of PBHFSP is given as follows:

**Indices:**
$i$: Index of jobs, $i = 1,2,\cdots,N$.
$j$: Index of stages, $j = 1,2,\cdots,S$.
$m$: Index of machines, $m = 1,2,\cdots,M$.
$t$: Index of the position on machines, $t = 1,2,\cdots,N$.

**Parameters:**
$N$: The number of jobs.
$S$: The number of stages.
$T_j$: Types of processing stages $j$. If stage $j$ is the parallel batch processing stage, $T_j$ is 1. If stage $j$ is discrete processing stage, $T_j$ is 0.
$M$: The number of machines.
$PT_{i,j,m}$: The processing time of job $i$ in stage $j$ on mach $m$.
$V_i$: The size of job $i$.
$MC_m$: The capacity of machine $m$.
$L$: A very large positive number.
$Ep$: The power consumption when the machine is loaded.
$Es$: The power consumption when the machine is idle.

**Variable:**
$ST_{i,j}$: The starting time of job $i$ in stage $j$.
$MST_{m,t}$: The starting time of the position $t$ in machine $m$.
$ENI_{m,t}$: The end time of job $i$ in stage $j$.
$ENP_{m,t}$: The end time of job $i$ in stage $j$.
$C_{max}$: the maximal completion time of all the factories.
$TEC$: total energy consumption.

**Decision variables:**
$x_{i,j,m,t} = \begin{cases} 1, \text{if } O_{i,j} \text{ processing on the position } t \text{ of the machine } m \\ 0, \text{otherwise} \end{cases}$

The objective function of DBHFSP:
$$minf_1 = C_{max}, minf_2 = TEC \quad (1)$$
subject to:
Equation (2) Constrain that a job can only be processed on one machine at a time.

$$\sum_m^M \sum_t^N x_{i,j,m,t} = 1, \forall i,j, PT_{i,j,m} > 0 \quad (2)$$

Equation (3) Constrain the order of processing of each operation. Limit the start time of each operation to exceed the completion time of the batch in the previous stage.

$$ST_{i,j} + \max_{\substack{1 \leq i1 \leq N, 1 \leq j1 \leq S-1 \\ 1 \leq m \leq M, 1 \leq t \leq N}} \{PT_{i,j,m} - L \times (2 - x_{i,j,m,t} - x_{i1,j1,m,t})\} \leq ST_{i,j+1}, \forall i,j, j < S, PT_{i1,j1,m} > 0 \quad (3)$$

The following formula constrains the batch size of the batch processing stage to not exceed the capacity of the machine it is on. Constrain the machine in the discrete processing stage to process only one job at a time.

$$\sum_{m}^{M}\sum_{t}^{N} x_{i,j,m,t} \times V_i \leq MC_m, \forall m, t, PT_{i,j,m} > 0, T_j = 1 \quad (4)$$

$$\sum_{m}^{M}\sum_{t}^{N} x_{i,j,m,t} \times V_i \leq 1, \forall m, t, PT_{i,j,m} > 0, T_j = 0 \quad (5)$$

Formula 6 constrains the batch processing time of two adjacent positions on each machine to not overlap.

$$MST_{m,t+1} - MST_{m,t} \geq \max_{1 \leq i \leq N, 1 \leq j \leq S-1}\{PT_{i,j,m} \times x_{i,j,m,t}\},$$
$$\forall m, t, t < N - 1 \quad (6)$$

The following formula constrains the start time of a job on a machine to be equal to the start time of the location of the machine on which it is located.

$$MST_{m,t} \geq ST_{i,j} - L \times (1 - x_{i,j,m,t}), \forall i, j, m, t, j < S - 1, PT_{i,j,m} > 0 \quad (7)$$

$$MST_{m,t} \leq ST_{i,j} + L \times (1 - x_{i,j,m,t}), \forall i, j, m, t, j < S - 1, PT_{i,j,m} > 0 \quad (8)$$

The following constraints the maximum completion time.

$$ST_{i,S-1} + \sum_{m}^{M}\sum_{t}^{N} x_{i,S-1,m,t} \times PT_{i,S-1,m} \leq C_{max}, \forall i \quad (9)$$

The following constraints the TEC.

$$ENI_{m,t} \geq \left(MST_{m,t+1} - MST_{m,t} - \max_{1 \leq i \leq N, 1 \leq j \leq S}\{PT_{i,j,m} \times x_{i,j,m,t}\}\right) \times Es, \forall m, t \quad (10)$$

$$ENP_{m,t} \geq \sum_{i}^{N}\sum_{j}^{S}\left(\max_{1 \leq i1 \leq N, 1 \leq j1 \leq S}\{PT_{i,j,m} - L \times (2 - x_{i,j,m,t} - x_{i1,j1,m,t})\}\right) \times Ep, \forall m, t, PT_{i1,j1,m} > 0 \quad (11)$$

$$TEC \geq ENP_{m,t} + ENI_{m,t}, \forall m, t \quad (12)$$

## IV. ALGORITHM DESCRIPTION

### A. The framework of the AMOEA/D

PBHFSP considers a parallel batch processing machine and simultaneously optimizes makespan and TEC, making the solution space more complex compared to FSP. Therefore, this paper proposes AMOEA/D to solve this problem. The AMOEA/D has the following critical parts. (1) a hybrid initialization strategy: Combining random initialization and heuristic rules based on knowledge of PBHFSP to generate a population with better diversity and convergence. (2) a critical-path-based neighborhood search: establishing a disjunctive graph model based on the knowledge of the PBHFSP. Optimizing two objectives based on the critical path, and adaptively adjusting the time based on learning experience from Q-learning and Decay Law. (3) population updates strategy: updated individuals re-match weight vectors and update weight vectors timely. The framework of AMOEA/D is shown in Algorithm 1.

### B. Encoding and Decoding

For scheduling problems with parallel batch processing stages, existing encoding methods are mostly based on job sequences, followed by decoding using heuristic rules. Heuristic rules such as batch first fit heuristic (BFF) and its derivative first fit large processing time heuristic (FFLPT) are utilized to generate feasible solutions for single-batch-stage processing machine scheduling problems[35]. However, these methods are not suitable for multi-stage batch processing scheduling problems. Wu et al.[36] employed an encoding method based on machine selection, followed by decoding using the first-come-first-served heuristic rule. Li et al.[37] adopted a segmented encoding method, where encoding based on machine and batch was used in batch processing stages, and the right-shifting heuristic was employed to allocate operations to batches. The above strategies all use heuristic rules to determine the batches where operations belong, which is advantageous for quickly obtaining a feasible solution but is not conducive to exploring the solution space.

Therefore, this article adopts the encoding method as shown in Fig. 1. Each machine has a processing linked list, and each element in the linked list is a batch. The arrangement order of batches is their processing order on the machine. The elements in the batch are the job numbers, as shown in the red box of 3 in the figure, indicating that the third operation of job 3 is processed on the second batch of machine 5. Due to the different machines available for each stage, the operation corresponding to the job number can be determined based on the machine where the job is located. Because this encoding can correspond to a complete solution, scheduling each operation one by one according to the definition and job processing order in the encoding can obtain fitness values.

| Algorithm 1. The framework of AMOEA/D |
|---|
| **Input**: $Popsize$, tabu length $L$, the size of neighborhood $T$, learning rate α, discount rate $r$ |
| **Output**: Nondominated solutions |
| 1. // Hybrid initialization |
| 2. $P: \{x_1, x_2, \cdots, x_{Popsize}\}$ ←initialization ($Popsize$); |
| 3. $\lambda': \{\lambda_1, \lambda_2, \cdots, \lambda_{Popsize}\}, [NM, \lambda]$ ← initialize weight vector ($Popsize$) and set neighbors for weight vectors; |
| 4. $[x, \lambda]$ ← matching weight vectors and individuals $(P, \lambda', T)$; |
| 5. Initial Q-table, ∅ ← $Taboo$ ; ∅ ← $ND$; |
| 6. $z^* \leftarrow \{\min_{1 \leq i \leq Popsize} f_1(x_i), \min_{1 \leq i \leq Popsize} f_2(x_i)\}$; |
| 7. **While** the termination criterion is not satisfied **do** |
| 8.     **For** $i = 1$ to $Popsize$ **do** |
| 9.         Choose a neighbor of $\lambda_i$ and find the solution $x_{\lambda_i}$ that maps to it; |
| 10.         $x$ ←Evolution operation ($x_{\lambda_i}$); |
| 11.         //Critical-path-based neighborhood search |
| 12.         $x'$ ←neighborhood search for makespan $(x, t)$; |
| 13.         Update $t$ according to Q-table; |
| 14.         Update Q-table; |
| 15.         $x''$ ←neighborhood search for TEC $(x')$; |
| 16.         //population updates |
| 17.         **For** $j = 1$ to $T$ **do** |
| 18.             Update neighborhood $(x'', \lambda_i,)$ |
| 19.         **End** |
| 20.     **End** |
| 21.     Update $\lambda'$; |
| 22.     Update $ND$; |
| 23. **End** |

## C. Hybrid Initialization

The randomly generated solution scheme can disperse the population, but due to the large solution space of PBHFSP, a completely random initialization strategy is difficult to generate superior solutions. Therefore, this algorithm adopts a mixed initialization strategy of random and heuristic rules, which not only preserves population diversity but also obtains high-quality solutions in the initialization phase. The hybrid initialization consists of two main steps: generating a job sequence, and scheduling according to the job sequence and heuristic rules based on knowledge. In Step 1, there are two rules for generating the job sequence: 1) randomly generating a job sequence, 2) dividing the jobs into multiple clusters based on the processing time using k-means algorithm, then randomly ordering each cluster to obtain the job sequence. Step 2 employs 5 heuristic rules:

1) MFBF: Prioritize selecting the machine that allows the earliest start time. If there is no batch on the machine, create a new batch; otherwise, check if the remaining volume of the last batch is less than the size of the job. If it is less, create a new batch on the machine. Finally, place the operation in the last batch on the machine.

2) MCBF: Schedule the operation on each machine in this phase, following the same batch selection rule as above, ultimately selecting the machine with the smallest completion time.

3) MTBF: Schedule the operation on each machine in this phase, following the same batch selection rule as above, ultimately selecting the machine with the smallest TEC.

4) MPBF: Select the machine with the minimum processing time for the operation, with the same batch allocation strategy as above.

5) MRBF: Randomly select a machine with the same batch allocation strategy as above.

In addition, after the completion of processing in the first phase, the subsequent job sequences are sorted in ascending order based on the release time of the jobs. Through combination, there are a total of 10 initialization strategies, with an equal number of individuals generated for each strategy in the initial population.

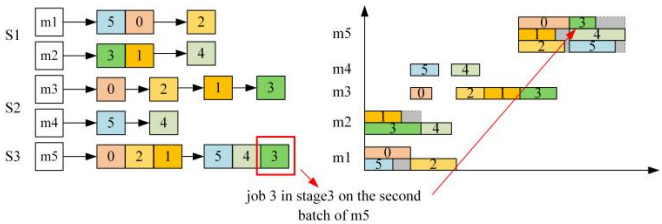

**Fig. 1.** The example of encoding and decoding.

## D. Critical-path based neighborhood search

### 1) Disjunctive Graph Representation of the PBHFSP

A schedule of HFSP can be represented by a disjunctive graph: $G := (N, A, E)$ [38]. The critical operations on the critical path in the disjunctive graph can optimize the schedule more accurately. In PBHFSP, considering the characteristic knowledge of simultaneous processing of operations within batches, this paper defines a disjunctive graph for associating batch processing stages and discrete processing stages. The disjunctive graph $G := (V, A, E, B)$ is defined as follows: $V$ is a collection of nodes for all operations $O_{i,j}$ of all jobs and includes two virtual nodes $s$ and $e$, representing the beginning and end of scheduling, respectively. Process arc $A$ is a set of directed arcs connecting consecutive operations of the same job. The directed arcs are used to link the operations processed sequentially on the same machine, and the set of these arcs is Machine arc $E$. For operations in parallel batch processing, the machine arc of the operation needs to be connected to each operation in its subsequent batch. Batch arc $B$ is a set of directed arcs that represent the relationship between batches of two adjacent stages of jobs. In detail, for an operation $O_{i,j}$ in the batch processing stage, first identify other operations within the same batch, then locate the batches in which these operations (including $O_{i,j-1}$, if $j \neq 0$) are performed in the previous stage, and finally establish arcs from these preceding stage operations to $O_{i,j}$. If the previous stage is a discrete processing stage, consider a single operation as a batch. Fig. 2 provides an example corresponding to the scheduling scheme in Fig. 1. Among them, nodes of the same color are on the same machine, and the set of black arcs is $A$, the set of blue arcs is represented by $E$, and the set of red arcs is $B$.

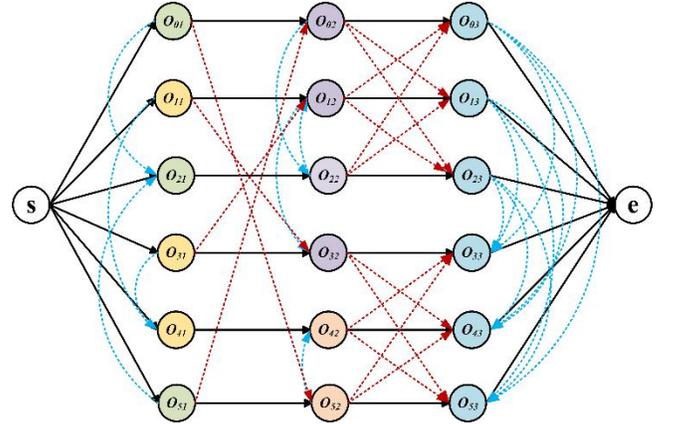

**Fig. 2.** The disjunctive graph representation of a scheduling.

In the disjunctive graph, the arc has no weight, but the weight $pt_v$ of each node $v$ represents the processing time of the operation on the machine. For node $v$, the set of adjacent nodes $Q = \{q_1, \cdots, q_k\}$ pointing to $v$ can be found in the disjunctive graph. $k$ is the inner degree of node $v$. The set of $v$ pointing to adjacent nodes $P = \{p_1, \cdots, p_l\}$ can be found in the disjunctive graph. $l$ is the outgoing degree of node $v$. The earliest and latest start times for each node are as follows:

$$Ve(v) = \max_{1 \leq i \leq k} \{Ve(q_i) + pt_{q_i}\} \quad (13)$$

$$Vl(v) = \min_{1 \leq i \leq k} \{Ve(p_i) - pt_v\} \quad (14)$$

### 2) Proof of Theorem

$L(u, v)$ represents the longest path from u to v, $L(s, e) = C_{max}$. Whether $f_v + p_v + t_v$ is equal to $C_{max}$ to determine whether a node v is on the critical path, where $f_v = L(s, v)$ and $t_v = L(v, e)$. In a scheduling, delete operation $v$ from the original machine (delete the machine-related arcs in $E$ and $B$ connected to the node $v$). The disjunctive graph corresponding to the changed scheduling is represented by $G^-$. This section proves the following three conclusions based on the approach of Monaldo[39].

a) *Theorem 1*: In $G^-$, reinserting the operation $v$ anywhere on the optional machine does not produce an infeasible solution.

*Proof:* $PJ_v$ is the preceding operation of operation $v$, while $SJ_v$ is the succeeding operation of $v$. The processing stage corresponding to $v$ is denoted by $j$. The set of all operations on machine k in $G^-$ is denoted by $OP_k$, where $x$ is an operation on $OP_k$. If there exists a path from $x$ to $v$ in $G^-$, then $v$ cannot be inserted into the batches before x; if there exists a path from $v$ to x in $G^-$, then $v$ cannot be inserted into batches after $x$.

First, in $G$, only Process arc in $A$ connected to operation $v$ exists as well as other operations in the same batch of $PJ_v$ point to the batch arc of $v$. Suppose there is a path from $x$ to $v$ in $G^-$, then there must be a path from $x$ to all operations in the batch where $PJ_v$ is located. However, since these operations belong to stage $j - 1$, there is no path from $x$ to these operations. Therefore, there is no path from $x$ to $v$ in $G^-$. Similarly, assuming there is a path from $x$ to $v$ in $G^-$, there must be a path from $SJ_v$ to $x$, but due to the constraints of processing order, there is no path from $v$ to $x$. Therefore, the theorem 1 can be proven.

b) *Theorem 2*: If $v$ is inserted into the batch where $u$ resides, and $u$ is the operation with the longest processing time in that batch. Selecting the batch to insert that with the smallest $BV = max(0, f_u^- - f_v^-) + max(0, p_{uk} - p_{vk}, t_u^- - t_v^-)$ under the condition of satisfying machine capacity constraints can minimize the value of $C_{max}$.

*Proof:* Insert operation $v$ into batch $B_u$. Add machine arcs and batch arcs for $v$ in $G^-$. The updated calculation method for $C_{max}^{B_u}$ is as follows:
$C_{max}^B = max(f_u^-, f_v^-) + max(p_{vk} + t_u^-, p_{uk} + t_u^-, p_{vk} + t_v^-)$
$= f_v^- + p_{vk} + t_v^- + max(0, f_u^- - f_v^-)$
$\quad + max(0, p_{uk} - p_{vk}, t_u^- - t_v^-)$
Therefore, when $max(0, f_u^- - f_v^-) + max(0, p_{uk} - p_{vk}, t_u^- - t_v^-)$ is minimized, $C_{max}^B$ can be minimized.

c) *Theorem 3*: Assuming that $B_{t-1}$ and $B_t$ are two adjacent batches on machine $k$. Select the operation with the smallest $RV = max(0, f_v^- - f_u) + t_v^- + max(0, p_{vk} - p_{uk})$ in $B_t$ and move it to $B_{t-1}$ to minimize $C_{max}^R$.

*Proof:* $v$ is the only operation on $B_{t-1}$, assuming that the capacity constraint of the machine is still satisfied after adding $u$ to $B_{t-1}$. After moving $u$ into $B_{t-1}$, the calculation of $C_{max}^R$ is as follows:

Due to
$$\begin{cases} f_u + p_u + t_u = f_v + p_v + t_v \\ f_u^- + p_u + t_u^- = f_v^- + p_v + t_v^- \end{cases}$$
and $f_u = f_u^-$,
so $f_v + t_v + t_u^- - t_u = f_v^- + t_v^-$
$\Rightarrow t_u^- = f_v^- + t_v^- - f_v - t_v + t_u$

$C_{max}^R = max(f_u^-, f_v^-) + max(p_{vk} + t_u^-, p_{uk} + t_u^-, p_{vk} + t_v^-)$
$= f_u^- + p_{uk} + max(0, f_v^- - f_u^-)$
$\quad + max(p_{vk} + t_u^- - p_{uk}, t_u^-, p_{vk} + t_v^- - p_{uk})$
$= f_u + p_{uk} + max(0, f_v^- - f_u)$
$\quad + max(p_{vk} + t_u^- - p_{uk}, t_u^-, p_{vk} + t_v^- - p_{uk})$
$= f_u + p_{uk} + t_u^- + max(0, f_v^- - f_u)$
$\quad + max(p_{vk} - p_{uk}, 0, p_{vk} - p_{uk} + t_v^- - t_u^-)$
$= f_u + f_u^- + p_{uk} + t_u - f_v - t_v - p_{uk} + p_{uk} + t_v^-$
$\quad + max(0, f_v^- - f_u) + max(p_{vk} - p_{uk}, 0, C_{max} - C_{max})$
$= f_u + C_{max} - f_v - t_v + t_v^-$
$\quad + max(0, f_v^- - f_u) + max(0, p_{vk} - p_{uk})$

Therefore, when $max(0, f_v^- - f_u) + max(0, p_{vk} - p_{uk})$ is minimized, $C_{max}^R$ can be minimized.

**Algorithm 2. Critical-batches based neighborhood search for makespan**

**Input**: an individual $x$, Iteration termination time $NeiTime = \theta \times S \times N \times M$;
**Output**: an individual $x_{nei}$
1. $t=0$; $x_{nei} \leftarrow x$;
2. **While** $t<NeiTime$ **do**
3.     **For** $i =1$ to $M$ **do**
4.         **Keyb** ←Identify the critical blocks of $x_{nei}$ on machine $i$.
5.         **For** $j =2$ to size of **Keyb** $- 1$ **do**
6.             $x'_{nei}$ ←Insert batch $j$ from **Keyb** into the block header;
7.             If $x'_{nei}$ dominates $x_{nei}$ then
8.                 $x_{nei} \leftarrow x'_{nei}$;
9.         **End**
10.    **End**
11. **End**
12. **While** $t<NeiTime$ **do**
13.    **KeyO** ←Identify the key operations of $x_{nei}$;
14.    **For** $i =1$ to $S$ **do**
15.         **KeyO** ←Randomly select an operation in stage $i$ from **KeyO**;
16.         $x^1_{nei}$ ←Batch insertion ($x^1_{nei}$,KeyO);
17.         If $x^1_{nei}$ dominates $x_{nei}$ then
18.             $x_{nei} \leftarrow x^1_{nei}$;
19.             Update **KeyO**;
20.    **End**
21.    **For** $i = 1$ to $S$ **do**
22.         **KeyO** ←Randomly select an operation in stage $i$ from **KeyO**;
23.         $x^2_{nei}$ ←Batch recombination ($x^2_{nei}$,KeyO);
24.         If $x^2_{nei}$ dominates $x_{nei}$ then
25.             $x_{nei} \leftarrow x^2_{nei}$;
26.             Update **KeyO**;
27.    **End**
28. **End**

*3) Critical-batches based neighborhood search for makespan*
Moving critical operations within the critical path can inevitably change the maximum completion time. Therefore, compared to random neighborhood search strategies, critical path-based neighborhood search can more quickly find better solutions. As show in Algorithm 2, this paper adopts three types of neighborhood structures: N6, batch insertion, and batch recombination. Since PBHFSP considers parallel batch processing, critical batches containing critical operations are also considered critical batches in the parallel batch processing stage. N6 involves searching for critical batches, while the other two involve searching for critical operations within batches. The specific strategies are as follows:

a) N6:

N6 was originally used in FJSP, where moving operations within critical blocks can lead to infeasible solutions. However, in the PBHFSP problem, according to Theorem 1, moving any critical operation within a critical block will not lead to infeasible solutions. Therefore, the operation of N6 is as follows: find all critical blocks based on the critical path, insert critical batches from the block into the block head, and insert critical batches from the block head into the block.

b) Batch insertion:

According to Theorem 2, remove critical operations from the machine first, then select the batch with the minimum $BV$ and satisfying the capacity constraint for insertion.

c) Batch recombination:

First, remove critical operations from the machine, then select two adjacent batches and insert the operations between them to form a new batch[39]. Next, select the operation with the minimum $RV$ from the batches following the new batch and insert it into the preceding batch. Repeat the above steps until the selected operation no longer satisfies the capacity constraint of machine, then change the batch forward and repeat the previous process. For more information on $RV$, please refer to Theorem 3. As shown in (b) of Fig. 3, inserting $J_5$ after $J_6$ reduces $C_{max}$.

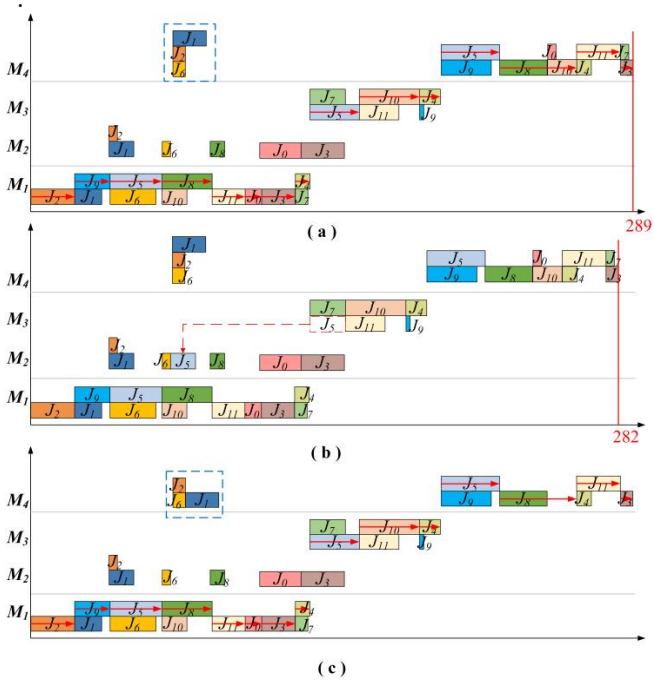

**Fig. 3.** The neighborhood search for a scheduling

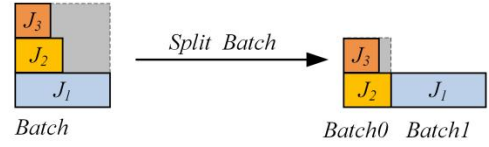

**Fig. 4.** The neighborhood search for TEC

*4) Noncritical-batches based neighborhood search for TEC*

The above neighborhood structure is mainly used to optimize makespan. Decreasing total processing energy consumption is also a key goal for PBHFSP. Reducing TEC can not only increase the profits of a company, but also reduce its burden on the environment. In PBHFSP, for a batch processed on a machine, the energy consumed during processing is the sum of the energy consumption of each operation in the batch. The energy consumption of each operation is the time it takes on the machine multiplied by the unit load power of the machine. Additionally, the power of the machine under load is greater than the power under no load. Therefore, a large difference in processing times among operations within the same batch will result in increased energy consumption. This section proposes an energy saving strategy based on batch splitting, which reduces TEC without increasing $C_{max}$. As shown in Fig. 4, assuming that the batch in the figure is a non-critical batch. The split batch will not cause an increase in $C_{max}$. The TEC of the scheduling scheme is the sum of the load energy consumption $Tp = \sum_{i=1}^{M} \sum_{j=1}^{BN} PT_j \times n_j \times Ep$ and the idle energy consumption $Ts = \sum_{i=1}^{M}(C_{max} - \sum_{j=1}^{BN} PT_j) \times Es$ of each machine. $BN$ is the number of batches on the machine, $PT_j$ is the processing time of the $j$ th batch, and $n_j$ is the number of operations in the batch. After batch decomposition, $Tp_{new} = Tp - (pt_1 - pt_2) \times Ep$, $Ts_{new} = Ts - pt_2 \times Es$. Therefore, the TEC after splitting is smaller.

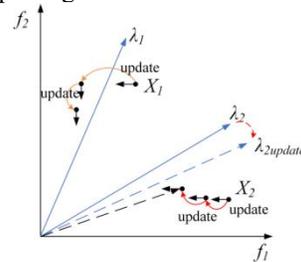

**Fig. 5.** Update weight vector

*5) Adaptation*

The parameter $NeiTime$ in the neighborhood search section can limit the scope of the neighborhood. Compared to using the same parameter for all data sizes, adaptive parameters can improve the search efficiency of the algorithm. According to the calculation formula of $NeiTime$, it varies with the size of the data. For problems with a large number of operations and machines, the solution space is also relatively large, thus requiring more time to search for better solutions. Among them, $\theta$ is a variable parameter determined based on the convergence of the current objective value. Therefore, this paper simulates the Decay Law and designs a method that adjusts the parameter $\theta$ adaptively based on iteration time. The curve of the Decay Law is also a curve that gradually converges over time. This method combines Q-learning to determine the time of neighborhood search based on the current iteration time, the state of the current solution, and the learning experience. The specific content of the adaptive strategy is as follows:

a) Actions and states:

$\theta$ has five levels {0.2, 0.3, 0.4, 0.5, 0.6}, and a value is selected from them during each neighborhood search. This process can be defined as an action, resulting in five actions in total. The purpose of neighborhood search is to find a better neighborhood solution and optimize each objective value. The current search state can be evaluated using $\Delta f1 = C_{max}^{nei} - C_{max}$ and $\Delta f2 = TEC_{nei} - TEC$. Therefore, four states can be combined: state 1: $\Delta f1 < 0, \Delta f2 \geq 0$; state 2: $\Delta f1 \geq 0$, $\Delta f2 < 0$; state 3: $\Delta f1 \geq 0$, $\Delta f2 \geq 0$; state 4: $\Delta f1 < 0, \Delta f2 < 0$.

b) Select action based on Decay Law:

When rand $> \varepsilon$, an action is randomly selected. Otherwise, the action with the maximum *Q-value* in the current state is chosen. Since the neighborhood search gradually converges over time, this part simulates the decay law by dynamically setting the value of ε based on the run time, as shown in (15).

$$\varepsilon = log\,(1/3)^{t/RunTime} \quad (15)$$

where t represents the current duration of the algorithm, while RunTime serves as the termination criterion for the algorithm, specifying the set running time for the algorithm.

c) Update Q-table: update the Q-table according to the following formula:

$$Q(s,a) = Q(s,a) + \alpha \times (R_s + r \times max_{a'}\,Q(s',a') - Q(s,a)) \quad (16)$$

where $\alpha$ is the learning rate, $r$ is the discount factor, $R_s$ is the reward value, s1 and s2 are 6, s3 is set to 0, and s4 is set to 10. In this context, $s'$ and $a'$ represent the current state and action, respectively

*E. Population updates*

The updating strategy in MOEA/D involves updating individuals based on the aggregated function value. In this algorithm, the Chebyshev aggregation method is employed, and the formula for the aggregated function is as follows:

$$g^{te}(x|\lambda,z^*) = max\{\lambda_1 \times (f_1 - z_1^*), \lambda_2 \times (f_2 - z_2^*)\} \quad (17)$$

In the traditional MOEA/D, when $g^{te}(x_{new}|\lambda_{nei},z^*) < g^{te}(x_{nei}|\lambda_{nei},z^*)$, all neighbors of an individual would be replaced with the $x_{new}$. This updating operation can quickly improve the quality of the population, but it lacks diversity. Especially in the later stages of iteration, there may be a small number of non-dominated solutions and uneven distribution of the Pareto front. Additionally, during the initial matching of weight vectors and individuals, there may be cases where the weight vector matched to an individual is not the most suitable, as there may be other solutions $x'$ with a smaller $g^{te}(x'|\lambda',z^*)$. Replacing these solutions with poor matching to the original weight vector during subsequent updates may result in the loss of promising solutions. Therefore, AMOEA/D only replaces the neighbor with the smallest $g^{te}(x_{new}|\lambda_{nei},z^*)$ when updating with $x_{new}$. To address the issue of optimal weight vector matching, when a new individual fails to replace the neighbors of $x$, all weight vectors are traversed to find the one that minimizes $g^{te}(x_{new}|\lambda'',z^*)$. If $g^{te}(x_{new}|\lambda'',z^*) < g^{te}(x''|\lambda'',z^*)$, the individual corresponding to that weight vector is replaced.

PBHFSP is a discontinuous problem, therefore the Pareto front of the problem may not be uniform, but the initial weight vectors are uniformly distributed. Consequently, some solutions may continuously update only one objective value. As shown in the Fig. 5, *X2* is constantly updated on the lower side of the weight vector, and as long as it does not cross the weight vector, the other objective value will not be optimized. This situation indicates that the true Pareto-optimal solution does not converge to this weight vector. Therefore, a parameter $L$ is set in this part. If a solution does not cross the weight vector after updating L times, the weight vector will be adjusted. The adjustment angle is half of the angle between the weight vector and its corresponding solution.

V. EXPERIMENTAL RESULTS AND ANALYSIS

*A. Instance setting*

In this section, we randomly generate 45 instances, where the number of jobs $N \in \{20,30,40,50,60\}$, the number of stages $S \in \{3,4,5\}$, Maximum number of machines in each stage $M_s \in \{3,4,5\}$. There is a 50% probability that any stage is a parallel batch processing stage. The capacity of the machine is randomly generated from a uniform discrete distribution between 10 and 15. The size of the job is randomly generated from a uniform discrete distribution between 1 and 10. The processing time is randomly generated from a uniform discrete distribution [1, 30]. All compared algorithms are coded in Java. All experiments are run on a Microsoft Windows 11 operating system with 20 GB RAM of memory and a 3.20 GHz Intel(R) Core (TM) i5-10505 CPU. We adopted three evaluation criteria, namely HV[40] and IGD[41]. Spread[42] illustrates the diversity of the solution set. HV is used to measure the overall performance of the algorithm, and the larger the HV is, the better the performance of the algorithm. The smaller the value of IGD, the better the performance of the algorithm. To ensure the fairness of the experiment, the running time of CPU was used as the termination condition. Each algorithm was run 10 times.

*B. Experimental Parameters*

The parameter settings have an impact on the performance of the algorithm. There are five parameters involved in the algorithm, and Taguchi's experimental method [43] is used to find the optimal combination of parameters in this section. The levels of each parameter are as follows: $Popsize = \{30,40,50,60\}$, $L = \{2,3,4,5\}$, $T = \{3,4,5,6\}$, $\alpha = \{0.1,0.2,0.3,0.4\}$, $r = \{0.95,0.90,0.85,0.8\}$. In this experiment, an orthogonal matrix $L_{16}(4^5)$ was used. The main effect plot for each metric is shown in the Fig. 6. Based on comprehensive evaluation, the optimal parameter configuration is as follows: $Popsize = 40$, $L = 2$, $T = 5$, $\alpha = 0.1$, r = 0.9.

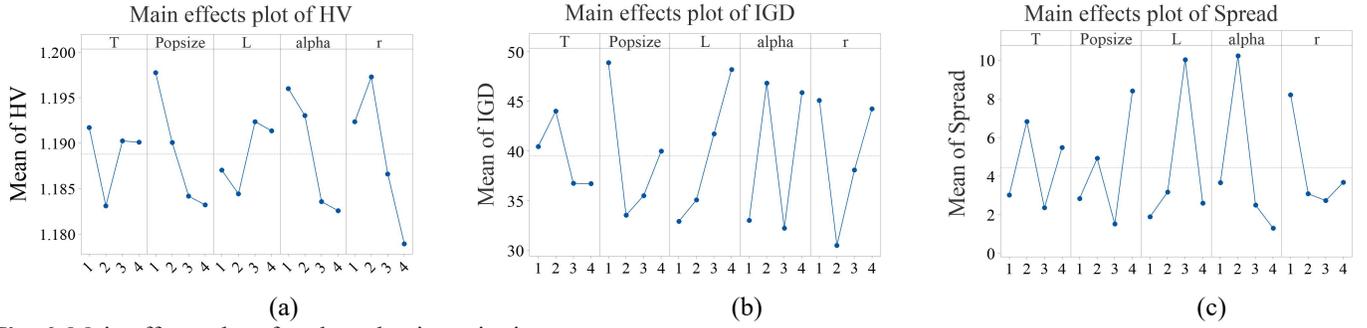

**Fig. 6.** Main effects plot of each evaluation criteria

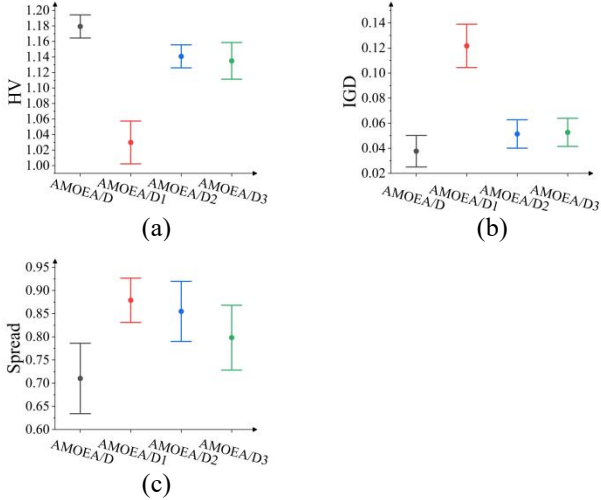

**Fig. 7.** The interval graph of each evaluation criteria for comparing each variant with AMOEA/D

## C. Effectiveness of Each Improvement Strategy of AMOEA/D

To verify the effectiveness of each improvement strategy in AMOEA/D, this experiment compared the performance of variants of AMOEA/D. Among them, AMOEA/D1 represents a variant of AMOEA/D that does not adopt the hybrid initialization strategy and adopts a random initialization rule; AMOEA/D2 represents a variant of AMOEA/D that does not adopt the critical path-based neighborhood search strategy proposed in this article; AMOEA/D3 represents a variant of AMOEA/D that adopts a general MOEA/D population update strategy. The experimental results on all instances are shown in Table S-1 (supplementary file). The interval graph based on the data in Table S-1 is shown in Fig. 7. In addition, we used the Friedman analysis of variance method to analyze the data of each group at the 0.05 level, and the analysis results are shown in Table I.

As shown in Fig. 7, overall, the AMOEA/D algorithm is the best. The solutions obtained by the initialization of AMOEA/D1 are not good enough, so it will take more time to search for solutions in the later stage. Therefore, the solutions obtained by the algorithm in the same running time are not good enough. The exploration capability of AMOEA/D2 is not sufficiently outstanding. Although it can obtain better solutions compared with AMOEA/D1, but AMOEA/D is closer to the real Pareto front than AMOEA/D2. As shown in the interval graph of spread in Fig. 7, comparing AMOEA/D and AMOEA/D3, AMOEA/D reduces the number of the same solutions in one iteration during population updating, which indeed enables the algorithm to find a more evenly distributed non-dominated solution set. The Friedman variance analysis is performed for all the values in Table S-1. The results are shown in Table I, and there are significant differences in the results obtained by each algorithm at the 0.05 level.

In conclusion, each strategy proposed in this paper has improved the performance of the algorithm. Specifically, AMOEA/D employs a hybrid initialization method, enabling the algorithm to obtain better solutions in the initial stage. By performing neighborhood search on these initial solutions, the algorithm can converge quickly. Additionally, the proposed updating strategy during population updating maintains the diversity of the population, resulting in a more uniform distribution of the final Pareto front.

## D. Comparisons to other algorithms

To further evaluate the effectiveness of the proposed AMOEA/D, it was compared to the current mainstream MODE[36], PACO[44], SDE[45] and C-NSGA-A[46]. Since this paper is the first study of PBHFSP, this section retains the part about parallel batch processing when reproducing comparative algorithms. Furthermore, some algorithms only have one batch processing phase, so heuristic rules based on first-come-first-served are used for discrete processing phases. The parameters in the algorithm are set according to the original text. The results of all algorithms are shown in Table S-2. Fig. 8 is an interval graph from the data in Table S-2, which more intuitively shows the comprehensive performance of each algorithm in solving all datasets. As shown in the figure, compared to other algorithms, AMOEA/D performs the best in solving PBHFSP. Due to space limitations, it is not possible to display the results of all datasets. Fig. 9 lists the Pareto fronts obtained by different algorithms for solving five datasets. The datasets are randomly selected from datasets corresponding to job numbers of different scales. As shown in the figure, the solutions obtained by AMOEA/D often dominate those of other algorithms. Furthermore, this section employs Friedman variance analysis to test the data in Table S-2, and the results are presented in Table II. There are significant differences in the results obtained by each algorithm at the 0.05 level.

In addition, the experiments in this section also investigate the performance of AMOEA/D on datasets of different sizes, as shown in Fig. 10. AMOEA/D performs best when solving problems with N=50. When N=20, the performance of AMOEA/D is similar to that of MODE. However, as the number of jobs increases, the advantage of AMOEA/D becomes more apparent compared to other algorithms. Furthermore, as the number of processing stages and the number of machines increases, the performance of AMOEA/D decreases. This is because more decisions need to be made during the solution process, resulting in a larger search space and requiring more search time for the algorithm. However, AMOEA/D outperformed the comparison algorithms when solving datasets of each size.

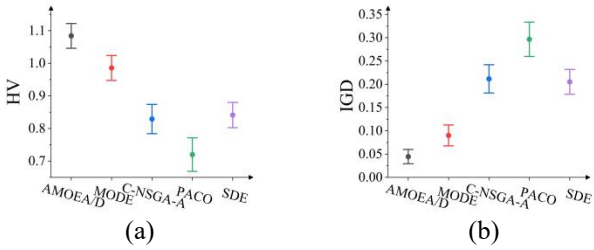

**Fig. 8.** The interval graph of HV for comparing each algorithm with AMOEA/D

TABLE I
FRIEDMAN ANALYSIS OF EACH VARIANTS OF AMOEA/D

| Algorithm | HV | | IGD | | Spread | |
|---|---|---|---|---|---|---|
| | Mean rank | p-value | Mean rank | p-value | Mean rank | p-value |
| AMOEA/D | **3.51111** | | **1.40000** | | **1.91111** | |
| AMOEA/D1 | 1.35556 | 9.30367E-14 | 3.66667 | 5.44009E-15 | 2.92222 | 0.00216 |
| AMOEA/D2 | 2.46667 | | 2.44444 | | 2.62222 | |
| AMOEA/D3 | 2.66667 | | 2.48889 | | 2.54444 | |

TABLE II
FRIEDMAN ANALYSIS OF EACH ALGORITH

| Algorithm | HV | | IGD | | Spread | |
|---|---|---|---|---|---|---|
| | Mean rank | p-value | Mean rank | p-value | Mean rank | p-value |
| AMOEA/D | **4.82222** | | **1.28889** | | **2.46667** | |
| MODE | 4.00000 | 1.03672E-21 | 1.97778 | 1.0244E-26 | 2.55556 | 0.0014 |
| PACO | 2.31111 | | 3.68889 | | 2.97778 | |
| SDE | 1.60000 | | 4.33333 | | 3.53333 | |
| C-NSGA-A | 2.26667 | | 3.71111 | | 3.46667 | |

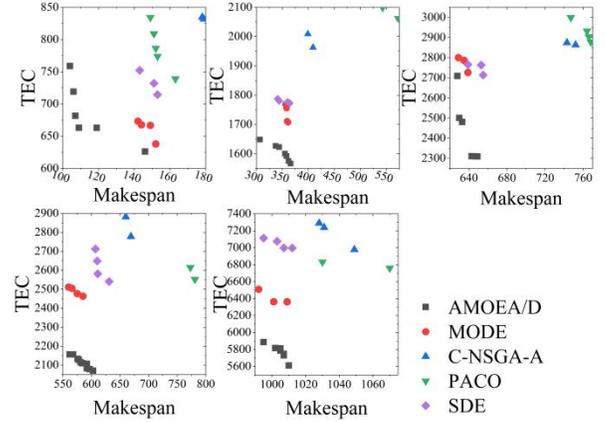

**Fig. 9.** Pareto Preface for each algorithm with different instances

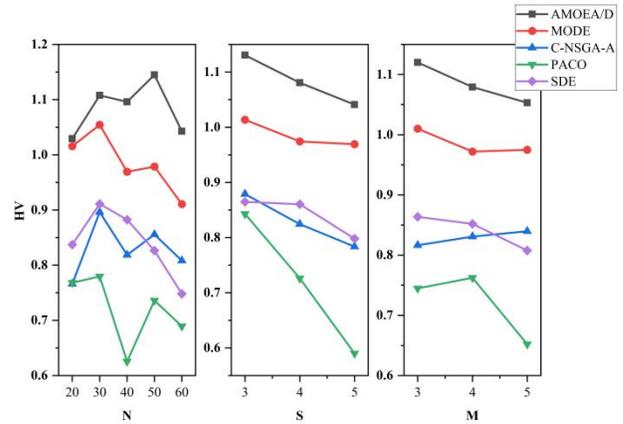

**Fig. 10.** The interaction plots between problem scale (N, S, M) and average HV values for AMOEA/D and comparative algorithms

Based on the corresponding HV and IGD of the algorithms, it is evident that the solutions obtained by AMOEA/D often dominate those of other algorithms. This is because the encoding rules and critical path-based search strategy adopted by AMOEA/D can more finely allocate each operation to suitable batches. However, most other algorithms use heuristic rules to assign jobs to their respective batches. Based on the Spread value corresponding to each algorithm, the improvement of AMOEA/D compared to MODE is not very significant. MODE employs heuristic rules for decoding and local search processes, leading to a faster iteration speed and allowing for a larger population size with better diversity. While AMOEA/D maintains population diversity through optimized population update strategies. Due to the significant time consumption of the domain search process, the population size for AMOEA/D cannot be set too large, as it would slow down the population iteration speed. Consequently, the improvement of AMOEA/D in solution distribution compared to MODE is not significant enough. However, in terms of overall performance, AMOEA/D has advantages over other algorithms in solving PBHFSP.

## VI. Conclusions

Multi-stage parallel batch processing scheduling is widely used in semiconductor manufacturing processes. Finding the optimal solution within a limited time for this problem is challenging. Therefore, this study proposes AMOEA/D to solve this problem while optimizing two objectives: maskespan and TEC. The algorithm adopts a hybrid initialization strategy based on knowledge of PBHFSP. The disjunctive graph model has been established based on knowledge of PBHFSP. Then, the algorithm designs a neighborhood structure based on the critical path. In addition, an improved population update strategy enhances the distribution of solutions. The effectiveness of each strategy is verified on multiple instances in this paper. Moreover, comparing AMOEA/D with several advanced algorithms, experiment results validate that AMOEA/D achieves superior comprehensive performance in solving PBHFSP compared to other algorithms.

In future research, it would be possible to further investigate neighborhood search strategies suitable for PBHFSP to improve search accuracy while reducing the required search time. Additionally, constraints on parallel batch processing machines can be adjusted based on actual production conditions to make the research problem more applicable to real-world production scenarios.

## References

[1] Fowler J W, Mönch L. "A survey of scheduling with parallel batch (p-batch) processing," European Journal Of Operational Research. 2022, 298(1): 1-24.
[2] Gonçalves J F, "Resende M G C. Biased random-key genetic algorithms for combinatorial optimization," Journal of Heuristics. 2011, 17(5): 487-525.
[3] Potts C N, Kovalyov M Y. "Scheduling with batching: A review," European Journal of Operational Research. 2000, 120(2): 228-249.
[4] Ozturk O, Espinouse M, Mascolo M D, et al. "Makespan minimisation on parallel batch processing machines with non-identical job sizes and release dates," International Journal of Production Research. 2012, 50(20): 6022-6035.
[5] Mathirajan M, Sivakumar A I. "A literature review, classification and simple meta-analysis on scheduling of batch processors in semiconductor," The International Journal of Advanced Manufacturing Technology. 2006, 29(9): 990-1001.
[6] Muter I. "Exact algorithms to minimize makespan on single and parallel batch processing machines," European Journal Of Operational Research. 2020, 285(2): 470-483.
[7] Bixby, Robert E. et al. "Short-Interval Detailed Production Scheduling in 300mm Semiconductor Manufacturing using Mixed Integer and Constraint Programming," in The 17th Annual SEMI/IEEE ASMC 2006 Conference 2006: 148-154.
[8] Hu K X, Che Y X, Ng T S, et al. "Unrelated parallel batch processing machine scheduling with time requirements and two-dimensional packing constraints," Computers & Operations Research. 2024, 162.
[9] Rocholl J, Mnch L, Fowler J. "Bi-criteria parallel batch machine scheduling to minimize total weighted tardiness and electricity cost," Zeitschrift fur Betriebswirtschaft. 2020(9): 90.
[10] Yu F, Lu C, Zhou J J, et al. "Mathematical model and knowledge-based iterated greedy algorithm for distributed assembly hybrid flow shop scheduling problem with dual-resource constraints," Expert Systems With Applications. 2024, 239.
[11] Zhou, Shengchao et al. "Distance matrix based heuristics to minimize makespan of parallel batch processing machines with arbitrary job sizes and release times," Appl. Soft Comput. 52 (2017): 630-641.
[12] Lu, Chao et al. " A Pareto-based collaborative multi-objective optimization algorithm for energy-efficient scheduling of distributed permutation flow-shop with limited buffers." Robotics Comput. Integr. Manuf. 74 (2022): 102277.
[13] Zhou S C, Liu M, Chen H P, et al. "An effective discrete differential evolution algorithm for scheduling uniform parallel batch processing machines with non-identical capacities and arbitrary job sizes," International Journal Of Production Economics. 2016, 179: 1-11.
[14] Wang Y, Jia Z H, Li K. "A multi-objective co-evolutionary algorithm of scheduling on parallel non-identical batch machines," Expert Systems With Applications. 2021, 167.
[15] Wu K, Huang E, Wang M C, et al. "Job scheduling of diffusion furnaces in semiconductor fabrication facilities," European Journal Of Operational Research. 2022, 301(1): 141-152.
[16] Beldar P, Moghtader M, Giret A, et al. "Non-identical parallel machines batch processing problem with release dates, due dates and variable maintenance activity to minimize total tardiness," Computers & Industrial Engineering. 2022, 168.
[17] Mecler D, Abu-Marrul V, Martinelli R, et al. "Iterated greedy algorithms for a complex parallel machine scheduling problem," European Journal of Operational Research. 2022, 300(2): 545-560.
[18] Xin X, Khan M I, Li S G. "Scheduling equal-length jobs with arbitrary sizes on uniform parallel batch machines," Open Mathematics. 2023, 21(1).
[19] Voß, Stefan and Andreas Witt. " Hybrid flow shop scheduling as a multi-mode multi-project scheduling problem with batching requirements: A real-world application, " International Journal of Production Economics 105 2007: 445-458.
[20] Amin-Naseri, Mohammad Reza and Mohammad Ali Beheshtinia. "Hybrid flow shop scheduling with parallel batching." International Journal of Production Economics 117 2009: 185-196.
[21] Rossi, Andrea, Andrea Pandolfi, and Michele Lanzetta. "Dynamic Set-up Rules for Hybrid Flow Shop Scheduling with Parallel Batching Machines." International Journal of Production Research. 2013, 52 (13): 3842–57. doi:10.1080/00207543.2013.835496.
[22] Costa, A., Cappadonna, F.A. & Fichera, S. "A novel genetic algorithm for the hybrid flow shop scheduling with parallel batching and eligibility constraints," Int J Adv Manuf Technol 75, 833–847 (2014).
[23] Ham A M, Cakici E. "Flexible job shop scheduling problem with parallel batch processing machines: MIP and CP approaches," Computers & Industrial Engineering. 2016, 102: 160-165.
[24] Ham A. "Flexible job shop scheduling problem for parallel batch processing machine with compatible job families," Applied Mathematical Modelling. 2017, 45: 551-562.
[25] Liu S W, Pei J, Cheng H, et al. "Two-stage hybrid flow shop scheduling on parallel batching machines considering a job-dependent deteriorating effect and non-identical job sizes," Applied Soft Computing. 2019, 84.
[26] Cheng L X, Tang Q H, Liu S L, et al. "Mathematical model and augmented simulated annealing algorithm for mixed-model assembly job shop scheduling problem with batch transfer," Knowledge-Based Systems. 2023, 279.
[27] Zeng C K, Qi G Q, Liu Z X, et al. "Auction-based approach with improved disjunctive graph model for job shop scheduling problem with parallel batch processing," Engineering Applications Of Artificial Intelligence. 2022, 110.
[28] Wang M, Zhang J, Zhang P, et al. "Independent double DQN-based multi-agent reinforcement learning approach for online two-stage hybrid flow shop scheduling with batch machines," Journal of Manufacturing Systems. 2022, 65: 694-708.
[29] Wu W, Hayashi T, Haruyasu K, et al. "Exact algorithms based on a constrained shortest path model for robust serial-batch and parallel-batch scheduling problems," European Journal of Operational Research. 2023, 307(1): 82-102.
[30] Wu X, Xie Z. "Improved MOEA/D with local search for solving multi-stage distributed reentrant hybrid flow shop scheduling problem," Expert Systems with Applications. 2023, 232: 120893.
[31] de Farias L, Araújo A. "A decomposition-based many-objective evolutionary algorithm updating weights when required," Swarm And Evolutionary Computation. 2022, 68.
[32] Shao W S, Shao Z S, Pi D C. "An Ant Colony Optimization Behavior-Based MOEA/D for Distributed Heterogeneous Hybrid Flow Shop Scheduling Problem Under Nonidentical Time-of-Use Electricity Tariffs," IEEE Transactions On Automation Science And Engineering. 2022, 19(4): 3379-3394.
[33] Liu Z G, Liang X, Hou L Y, et al. "Multi-Strategy Dynamic Evolution-Based Improved MOEA/D Algorithm for Solving Multi-Objective

Fuzzy Flexible Job Shop Scheduling Problem," IEEE Access. 2023, 11: 54596-54606.

[34] Cao J, Zhang J L, Zhao F Q, et al. "A two-stage evolutionary strategy based MOEA/D to multi-objective problems," Expert Systems With Applications. 2021, 185.

[35] Uzsoy R. "Scheduling a single batch processing machine with non-identical job sizes," International Journal of Production Research. 1994, 32(7): 1615-1635.

[36] Wu X L, Yuan Q, Wang L. "Multiobjective Differential Evolution Algorithm for Solving Robotic Cell Scheduling Problem With Batch-Processing Machines," IEEE Transactions On Automation Science And Engineering. 2021, 18(2): 757-775.

[37] Li, Jun-qing et al. "Hybrid Artificial Bee Colony Algorithm for a Parallel Batching Distributed Flow-Shop Problem With Deteriorating Jobs," Quantum electronics. 2020, 50(6).

[38] Fan J X, Li Y L, Xie J, et al. "A Hybrid Evolutionary Algorithm Using Two Solution Representations for Hybrid Flow-Shop Scheduling Problem," IEEE Transactions On Cybernetics. 2023, 53(3): 1752-1764.

[39] Mastrolilli M, Gambardella L M. "Effective Neighbourhood Functions for the Flexible Job Shop Problem," Journal of Scheduling. 2015, 3(1): 3-20.

[40] While L, Hingston P, Barone L, et al. "A faster algorithm for calculating hyper volume," IEEE Transactions On Evolutionary Computation. 2006, 10(1): 29-38.

[41] Zitzler E, Deb K, Thiele L. "Comparison of Multiobjective Evolutionary Algorithms: Empirical Results," Evolutionary Computation. 2000, 8(2): 173-195.

[42] Deb K, Pratap A, Agarwal S, et al. "A fast and elitist multiobjective genetic algorithm: NSGA-II," IEEE Transactions On Evolutionary Computation. 2002, 6(2): 182-197.

[43] Nostrand R C V. "Design of Experiments Using the Taguchi Approach: 16 Steps to Product and Process Improvement," Technometrics. 2001, 44(3): 289.

[44] Jia Z H, Zhang Y L, Leung J, et al. "Bi-criteria ant colony optimization algorithm for minimizing makespan and energy consumption on parallel batch machines," Applied Soft Computing. 2017, 55: 226-237.

[45] Zhou S C, Xing L N, Zheng X, et al. "A Self-Adaptive Differential Evolution Algorithm for Scheduling a Single Batch-Processing Machine With Arbitrary Job Sizes and Release Times," IEEE Transactions On Cybernetics. 2021, 51(3): 1430-1442.

[46] Li K, Zhang H, Chu C B, et al. "A bi-objective evolutionary algorithm scheduled on uniform parallel batch processing machines," Expert Systems With Applications. 2022, 204.